\documentclass[10pt,twocolumn,letterpaper]{article}

\usepackage{cvpr}
\usepackage{times}
\usepackage{epsfig}
\usepackage{graphicx}
\usepackage{amsmath}
\usepackage{amssymb}

\usepackage{acronym}
\usepackage{multirow}
\usepackage{dblfloatfix}
\usepackage{subfig}
\usepackage{float}
\usepackage{stackengine}

\graphicspath{{Images/}}

\acrodef{SLAM}{Simultaneous Localization and Mapping}
\acrodef{VO}{Visual Odometry}
\acrodef{VLAD}{Vector of Locally Aggregated Descriptors}
\acrodef{FCN}{Fully Convolutional Network}
\acrodef{SPP}{Spatial Pooling Pyramid}
\acrodef{SAND}{Scale-Adaptive Neural Dense}
\acrodef{AUC}{Area Under the Curve}
\acrodef{DVF}{Deja-Vu feature}

\def\p#1{\boldsymbol{p}_#1} 
\def\F#1{\boldsymbol{F}_#1} 
\def\I#1{\boldsymbol{I}_#1} 
\def\codelink{{\small \url{https://github.com/jspenmar/DejaVu_Features}}}

\usepackage[breaklinks=true,bookmarks=false]{hyperref}

\cvprfinalcopy 

\def\cvprPaperID{303} 

\ifcvprfinal\fi
\begin{document}

\title{Same Features, Different Day: Weakly Supervised Feature Learning for Seasonal Invariance}

\author{Jaime Spencer,\quad Richard Bowden,\quad Simon Hadfield\\
Centre for Vision, Speech and Signal Processing (CVSSP)\\
University of Surrey\\
{\tt\small \{jaime.spencer, r.bowden, s.hadfield\}@surrey.ac.uk}}

\maketitle
\thispagestyle{empty}

\begin{abstract}
``Like night and day'' is a commonly used expression to imply that two things are completely different. Unfortunately, this tends to be the case for current visual feature representations of the same scene across varying seasons or times of day. The aim of this paper is to provide a dense feature representation that can be used to perform localization, sparse matching or image retrieval, regardless of the current seasonal or temporal appearance.
\looseness=-1

Recently, there have been several proposed methodologies for deep learning dense feature representations. These methods make use of ground truth pixel-wise correspondences between pairs of images and  focus on the spatial properties of the features. As such, they don't address temporal or seasonal variation. Furthermore, obtaining the required pixel-wise correspondence data to train in cross-seasonal environments is highly complex in most scenarios. 
\looseness=-1

We propose Deja-Vu, a weakly supervised approach to learning season invariant features that does not require pixel-wise ground truth data. The proposed system only requires coarse labels indicating if two images correspond to the same location or not. From these labels, the network is trained to produce ``similar'' dense feature maps for corresponding locations despite environmental changes. Code will be made available at: \codelink
\end{abstract}

\section{Introduction}
\begin{figure}[t!]
\centering
\subfloat[Traditional]{\includegraphics[width=1.0\linewidth]{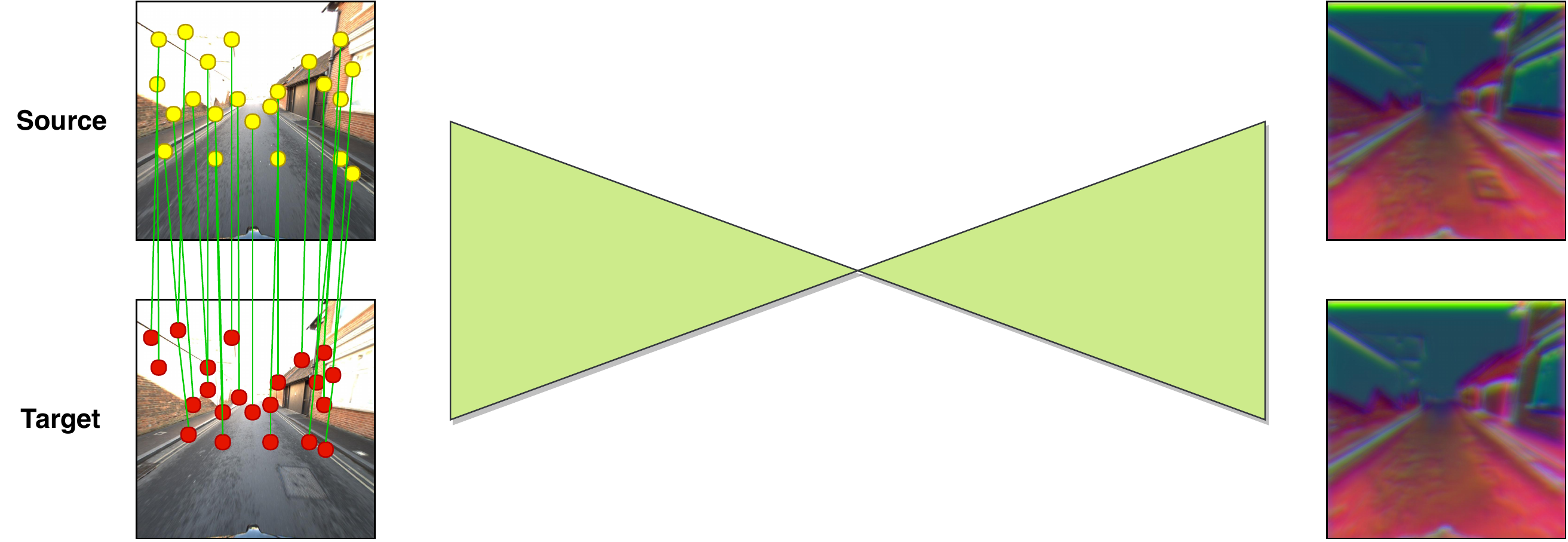}} \\
\subfloat[Proposed]{\includegraphics[width=1.0\linewidth]{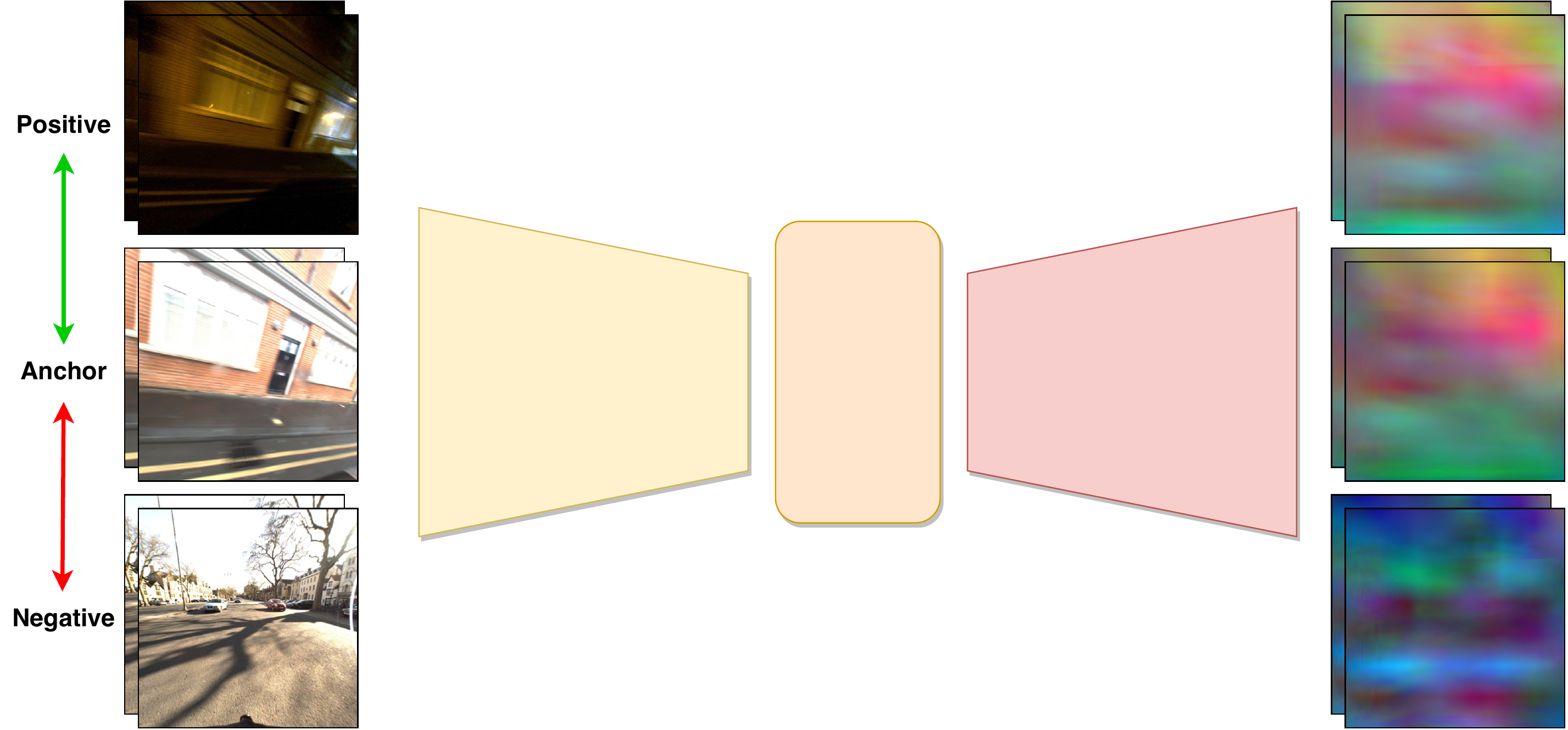}}
\caption{Traditional methods for learning dense features require pixel-wise correspondences, making cross-seasonal training nearly impossible. We propose a novel method for dense feature training requiring only image level correspondences and relational labels.}
\vspace*{-0.3cm}
\label{fig: intro}
\end{figure}

Feature extraction and representation is a core component of computer vision.
In this paper we propose a novel approach to feature learning with applications in a multitude of tasks. 
In particular, this paper addresses the highly challenging task of learning features which are robust to temporal appearance changes. 
This includes both short-term and long-term changes, \eg day vs. night and summer vs. winter, respectively.
This is important in scenarios such as autonomous driving, where the vehicle must be capable of operating reliably regardless of the current season or weather. 

Traditional hand-crafted features, such as SIFT \cite{Lowe2004} and ORB \cite{Rublee2011}, typically fail to obtain reliable matches in cross-domain environments since they haven't been designed to handle these changes.
More recently, there have been several deep learning techniques proposed \cite{Schmidt2017, Schuster2019, Spencer2019} to learn dense feature representations.
These methods tend to use a set of pixel-wise correspondences to obtain relational labels indicating similarity or dissimilarity between different image regions.
As such, these techniques focus on the spatial properties of the learned features.

However, none of these methods address the huge visual appearance variation that results from longer temporal windows.
This is likely due to the heavy biases in the commonly used training datasets \cite{Cordts2016, Geiger2012, Huang2018}, which do not incorporate seasonal variation. 
A limiting factor to this is acquiring ground truth correspondences for training.
Even if the dataset does have data across multiple seasons \cite{Maddern}, obtaining the pixel-wise ground truth correspondences required for these techniques is non-trivial.
The noise from GPS and drift from \ac{VO} make pointcloud alignment unreliable and, by the very definition of the problem, appearance cannot be used to solve cross-seasonal correspondence.

In order to overcome this, we instead opt for a weakly supervised approach. 
Rather than obtaining relational labels at the pixel level, we use coarse labels indicating if two images were taken at the same location. 
The network is then trained to produce globally ``\emph{similar}" dense feature maps for corresponding locations. 
An illustration of this process can be found in Figure \ref{fig: intro}.
This allows us to obtain large amounts of training data without requiring pixel-wise cross-seasonal alignment.
This paper introduces one of the only approaches capable of using holistic image-level correspondence as ground truth to supervise dense pixel-wise feature learning.

The remainder of this paper describes the details of the proposed approach. 
This includes the architecture of the \ac{DVF} network and the similarity metric used to train it.
We show the properties of the learned features, demonstrating their seasonal invariance.
Finally, we discuss the potential applications of these features, most notably in areas such as self-localization.

The main contributions can be summarized as follows:
\begin{enumerate}
	\item We propose a novel dense feature learning framework focused on invariance to seasonal and visio-temporal changes.
	\item We achieve this in a weakly supervised manner, requiring only rough cross-seasonal image alignment rather than pixel-level correspondences and yet we solve the pixel-level feature description problem.
	\item Finally, we propose a novel method for performing localization based on the aforementioned similarity metric, which makes full use of the dense feature maps.
\end{enumerate}  
\begin{figure*}[th!]
\vspace*{-1cm}
\centering
\includegraphics[width=\linewidth]{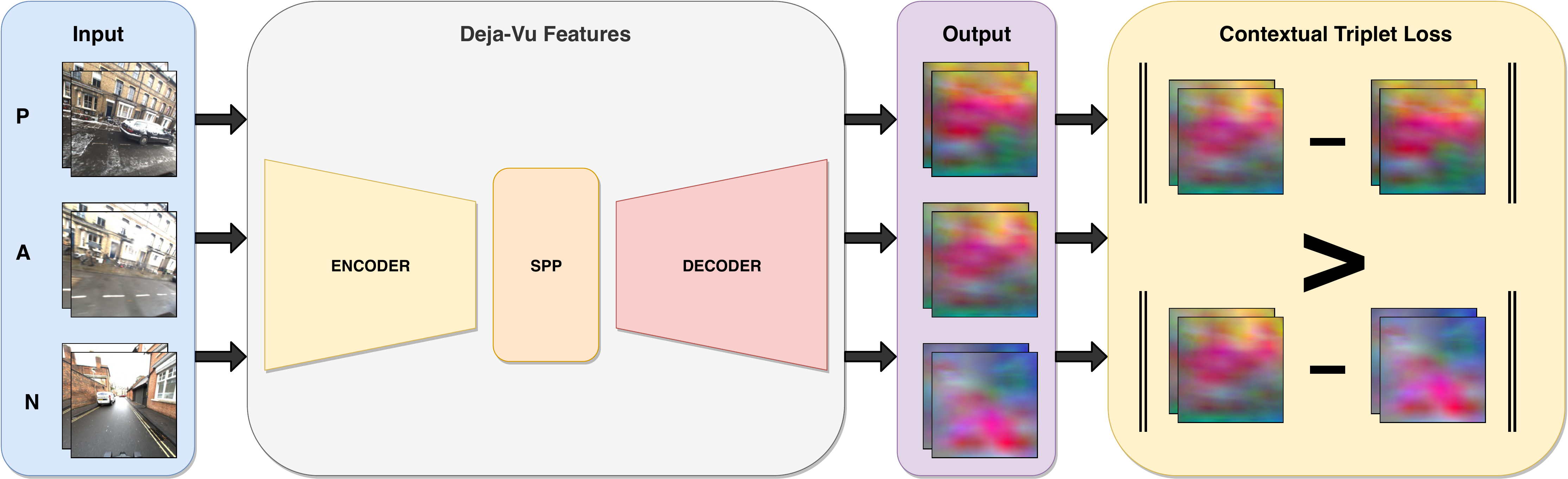}
\caption{Proposed methodology overview. The network is trained with consecutive \textbf{A}nchor - \textbf{P}ositive - \textbf{N}egative triplets corresponding to images of the same or different locations, respectively. The similarity metric doesn't require spatial alignment between images or perfect pixel-wise correspondences. }
\vspace*{-0.3cm}
\label{fig: net}
\end{figure*}

\section{Related Work}
Historically, hand-crafted sparse features have been widely popular \cite{Tuytelaars2008}. 
Notable examples include SIFT \cite{Lowe2004} and ORB \cite{Rublee2012}.
These continue to be used in current applications for \ac{SLAM} \cite{Mur-Artal2015a, Mur-Artal2016} and \ac{VO} estimation \cite{Zhou2017}.
Meanwhile, Li \etal \cite{Li2012} and Sattler \etal \cite{Sattler} use SIFT descriptors in a 3D pointcloud to perform global localization.  
On the other hand, Krajnik \etal \cite{Krajnik2015, Krajnik2017} introduced GRIEF, based on an evolutionary algorithm to refine BRIEF \cite{Calonder2010} comparisons. 
These features were subsequently applied to relocalization in \cite{Krajnik2017a}.
LIFT \cite{MooYi} and LF-Net \cite{Ono} instead train a sequential pipeline of networks in order to learn keypoint detection, orientation estimation and feature description.
However, Valgren and Lilienthal \cite{Valgren2010} demonstrate how the performance of sparse features degrades as the seasonal variation increases. Stylianou \etal \cite{Stylianou2015} instead claim that changes in illumination and keypoint detection failures are the main degrading factors.

Alternative approaches aggregate sparse features to form super-pixel image representations.
Such is the case in the work of Neubert \etal \cite{Neubert2013, Neubert2015} and Naseer \etal \cite{Naseer2014}, who aggregate SURF \cite{Bay} and HOG \cite{Dalal} features, respectively.
Other methods take this idea further and learn how to combine sparse features into a single holistic image descriptor.
Some of the most notable examples include the Bag of Words \cite{Csurka} and the Fisher Kernel \cite{Jaakkola, VanDerMaaten2011}.
Such methods have been applied to localization and mapping in \cite{Mousavian, Filliat2007}.
As an extension to these methods, Jegou \etal propose the \ac{VLAD} \cite{Jegou2010}, simplifying the computational complexity whilst maintaining performance. 
Torii \etal introduced DenseVLAD \cite{Torii2018}, combining RootSIFT \cite{Arandjelovic2012} and view synthesis to perform localization. 
Meanwhile, the contextual loss \cite{Mechrez2018a} has been proposed as a metric for similarity between non-aligned images. 
However, it has never been used in the context of explicit feature learning.

Since the rise of deep learning, methods have focused on the aggregation of intermediate pretrained features \cite{Naseer2015, Neubert2016}.
Xia \etal \cite{Xia2016} incorporated PCANet \cite{Chan} into a \ac{SLAM} loop closure system.
Meanwhile, \ac{VLAD} was adapted into deep learning frameworks such as NetVLAD \cite{Arandjelovi} and directly applied to place localization. 
Other approaches to holistic image feature learning include  \cite{Chen2017, Ivanovic}.
These methods make use of relational labels indicating similarity or dissimilarity to train their networks.
As such, they rely on losses such as contrastive \cite{Hadsell2006} and triplet \cite{Schroff2015a} loss.

These deep learning methods focus on producing a single descriptor representing the whole image.
However, Sattler \etal \cite{Sattler2018a} conclude that in order to solve complex localization problems it is necessary to learn dense feature descriptors.
Dusmanu \etal \cite{Dusmanu2019} opt for a ``describe-then-detect'' approach, where non-maximal-suppression is used to detect keypoints of interest in a dense feature map.
Meanwhile, Schuster \etal \cite{Schuster2019} introduce SDC-Net, focused on the design of an architecture based on the use of stacked dilated convolutions.
Schmidt \etal \cite{Schmidt2017} introduce a pixel-wise version of the contrastive loss used to train a network to produce dense matches between DynamicFusion \cite{Newcombea} and KinectFusion \cite{Izadi2011} models.
Fathy \etal \cite{Fathy2018a} employ a combination of losses in order to train coarse-to-fine dense feature descriptors.
Spencer \etal \cite{Spencer2019} extended these methods to introduce a more generic concept of scale through spatial negative mining.
The main drawback of these methods is that they do not tackle seasonal invariance.

In this paper we propose a new framework to learn dense seasonal invariant representations.
This is done in a largely unsupervised manner, greatly expanding the use cases of this feature learning framework. 
Furthermore, we extend contextual loss to create a relational loss based on a triplet configuration.

\section{Deja-Vu Features}
The aim of this work is to provide a dense feature descriptor representation for a given image.
This representation must be capable of describing features uniquely such that short-term feature matching is possible, but with sufficient invariance to temporal appearance variation such that features can also be matched between day \& night or winter \& summer.

An overview of the proposed methodology can be found in Figure \ref{fig: net}.
At the core of the system lies a \ac{FCN} formed from residual blocks and skip connections.
By utilizing only convolutions, the network is not restricted to a specific input size and allows for the estimation of a feature at every pixel in the image.
The final stage of the encoder makes use of a \ac{SPP} block, with average pooling branches of size 32, 16, 8 and 4, respectively. 
This allows the network to incorporate information from various scales and provide a more detailed representation of the input. 

Formally, we define Deja-Vu to produce a dense $n$-dimensional representation at every pixel in the input image, $\boldsymbol{F} \in \mathbb{R}^{H \times W \times n}$, obtained by
\vspace{-0.2cm}
\begin{equation} \label{eq: network}
	\boldsymbol{F}  = \Phi(\boldsymbol{I} | w),
	\vspace{-0.2cm}
\end{equation}
where $\boldsymbol{I} \in \mathbb{N}^{H \times W \times 3}$ is the corresponding input image and $\Phi$ a network parametrized by a set of weights $w$. 

\subsection{Contextual Similarity}
\begin{figure*}[!h]
\centering
\includegraphics[width=\linewidth]{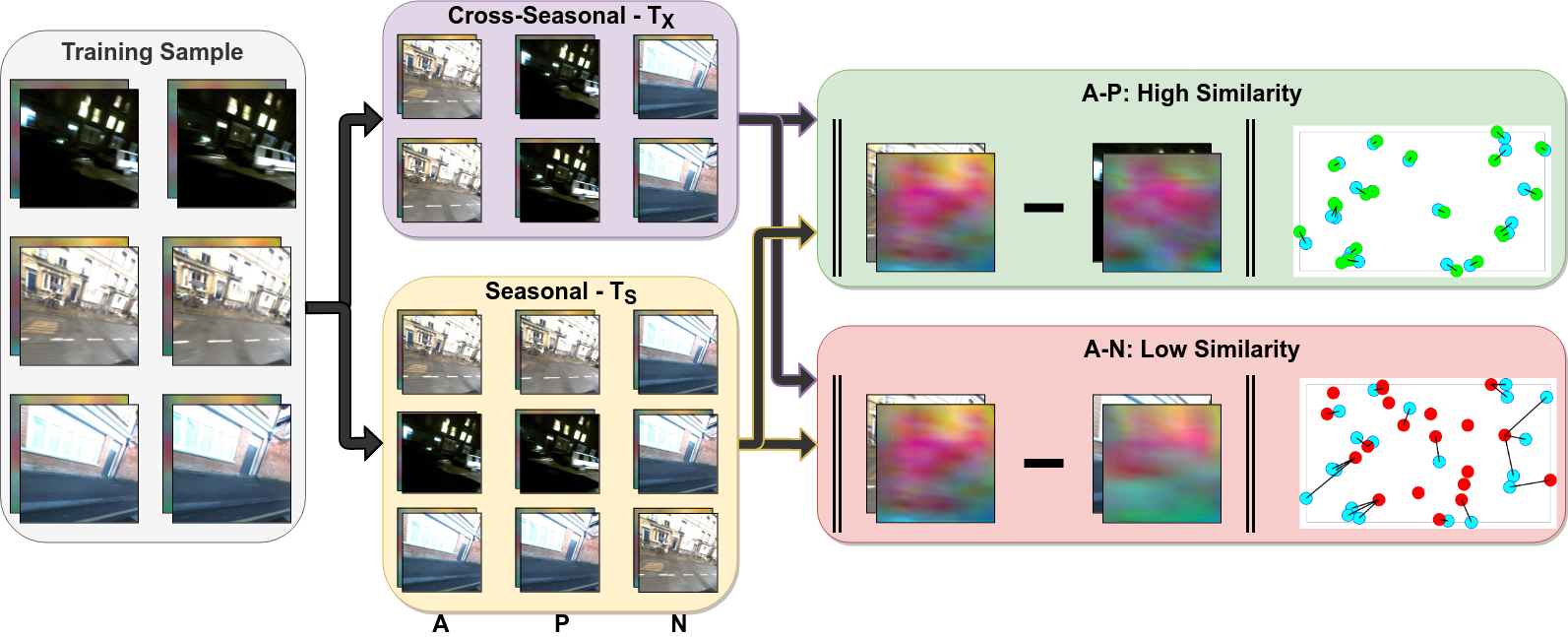}
\caption{Contextual triplet loss framework. The triplets formed by each training sample can be divided into seasonal or cross-seasonal triplets, each contributing to the short-term and long-term matching performance. In the case of positive pairs, each feature in the image should only match to a single feature in the other image.}
\label{fig: triplet}
\end{figure*}

In order to determine the similarity between two images, $\I1$ \& $\I2$, we first obtain their features, $\F1$ \& $\F2$, using (\ref{eq: network}).
We then take inspiration from \cite{Mechrez2018a} to quantify how uniquely each feature in $\F1$ matches to a single location in $\F2$.
This allows us to compare feature maps of the same location without requiring pixel-wise matches or perfect photometric alignment. 
In the context of localization and revisitation, this means that two images of the same location should be regarded as similar, whereas any other pair should be dissimilar. 

To formalize this idea, two images are considered similar if each feature descriptor in $\I1$ has a matching feature in $\I2$ that is significantly closer in the embedded feature space than any other features in that image.
Given a single feature at a 2D point $\p1$, its set of distances with respect to the other feature map is defined as
\begin{equation}
	D(\p1) = \{ \; ||\F1(\p1) - \F2(\p2)|| \; : \forall \; \p2 \; \}.
\end{equation}
We then normalize this set of distances according to 
\begin{equation}
	\widetilde{D}(\p1) = \frac{D(\p1)}{\min(D(\p1)) + \epsilon},
\end{equation}
where $\epsilon = 1e-5$.
Intuitively, this is similar to performing a traditional ratio test on the feature distances. 
In general, the best match will have $\widetilde{D} = 1$.
The rest of the points are then described as the ratio with respect to the best match in the range $\widetilde{D} = [1, \infty)$.
The set of normalized similarities between $\p1$ and all of $\F2$ is then given using a softmax function
\begin{equation}
	S(\p1) = exp \left( \frac{1 - \widetilde{D}(\p1)}{h} \right),
\end{equation}
\begin{equation}
	\widetilde{S}(\p1) = \frac{S(\p1)}{\sum_{\p2}{S(\p2)}},
\end{equation}
where $h$ represents the band-width parameter controlling the ``hardness'' of the similarity margin.
In this case, the best match results in a value of $S = 1$ and $S$ will tend to 0 for large values of $\widetilde{D}$.
$\widetilde{S}$ is therefore maximised by a single low and many high values of $\widetilde{D}$, \ie cases where there is a unique match.

Following these definitions, we can now represent the global similarity between the original pair of images as 
\begin{equation} \label{eq: cx_sim}
	CX(\F1, \F2) = \frac{1}{N} \sum_{\p1} \max \widetilde{S}(\p1),
\end{equation}
where $N$ is the total number of features $\p1$.
Since this is an average of the normalised pixel-wise similarities, the resulting metric is constrained to the range [0, 1], indicating completely different or identical feature maps, respectively.
As such, this encodes both the distances and uniqueness of the feature space without enforcing spatial constraints.
This similarity metric can now be used at inference time to determine if two feature maps are likely to represent the same location.

\subsection{Contextual Triplet Loss}
Since we make use of relational labels between images, \ie if the images correspond to approximately the same location or not, the similarity metric is introduced into a triplet loss framework.
In a traditional triplet loss, the aim is to minimize positive feature embedding distances, \textbf{AP}, and separate them from negative pairs \textbf{AN} by at least a set margin $m$. \looseness=-1

However, in the context of relocalization, positive pairs should be those with a high similarity.
We therefore introduce a modified triplet loss inspired by (\ref{eq: cx_sim}) to take this into account:
\vspace{1cm}
\begin{equation}
	T = \{ \I{A}, \I{P}, \I{N} \}, 
\end{equation}
\begin{equation}
	l(T)  = \max(CX(\F{A}, \F{N}) - CX(\F{A}, \F{P}) + m, 0).
\end{equation}
Given an \textbf{A}nchor image, the \textbf{P}ositive sample is obtained from the same location in a different weather/season.
On the other hand, the \textbf{N}egative corresponds to an image from a different location and any season.

Each training sample is composed of two consecutive frames of triplets. 
This framework allows us to introduce additional triplets, which help to provide additional consistency within each season and aid short-term matching.
This results in a total of five triplets per training sample, illustrated in Figure \ref{fig: triplet}.
In order to incorporate the information from all the triplets, the final loss is defined as
\begin{equation}
	L = \frac{1}{N_X} \sum_{T_X} l(T_X) 
	+ \frac{\alpha}{N_S} \sum_{T_S} l(T_S), 
\end{equation}
where $T_X$ and $T_S$ are the sets of seasonal and cross-seasonal triplets, $N_X$ and $N_S$ are the respective number of triplets in each category and $\alpha$ is a balancing weight in the range $[0, 1]$.
Once again, the image-level labels of \textbf{A}, \textbf{P} \&  \textbf{N} are used to drive pixel-wise feature training.

\section{Results}
\begin{figure*}
\centering
\hspace{-0.1cm}
\subfloat{\includegraphics[width=0.15\linewidth]{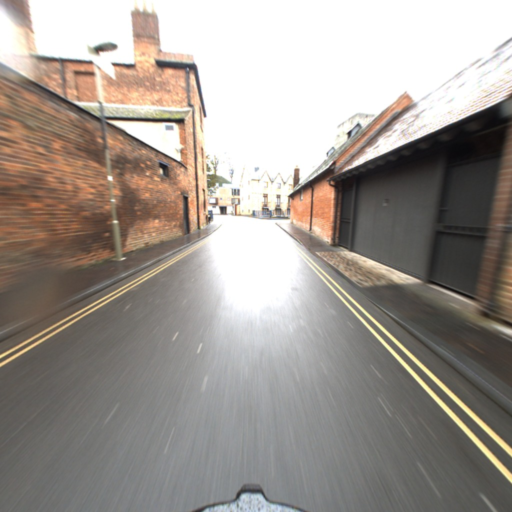}} \hspace{0.05cm}
\subfloat{\includegraphics[width=0.15\linewidth]{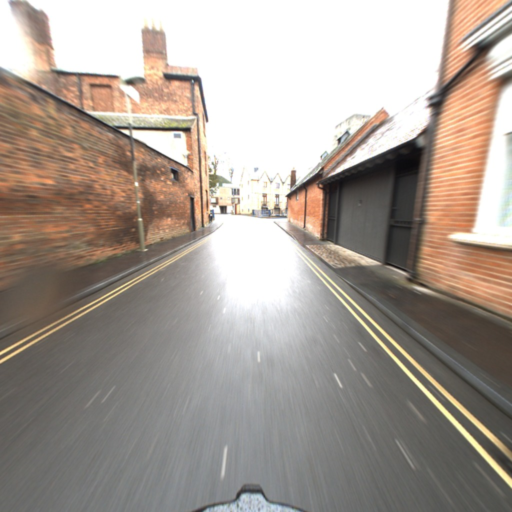}} \hspace{0.05cm}
\subfloat{\includegraphics[width=0.15\linewidth]{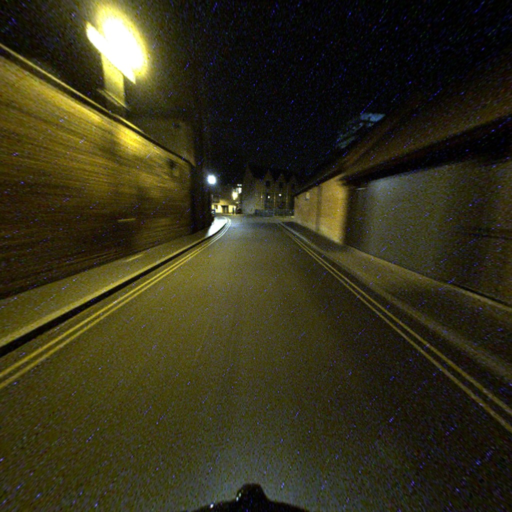}} \hspace{0.05cm}
\subfloat{\includegraphics[width=0.15\linewidth]{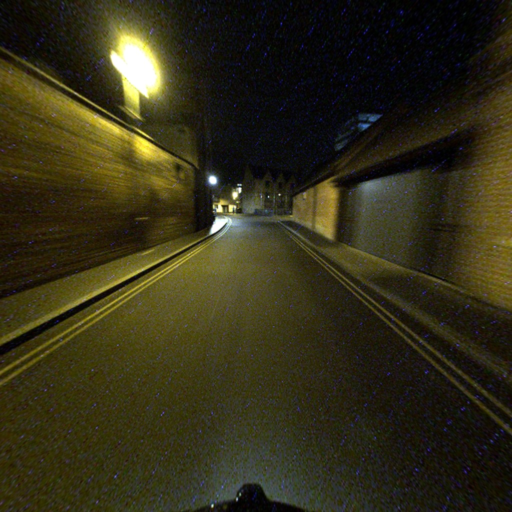}} \hspace{0.05cm}
\subfloat{\includegraphics[width=0.15\linewidth]{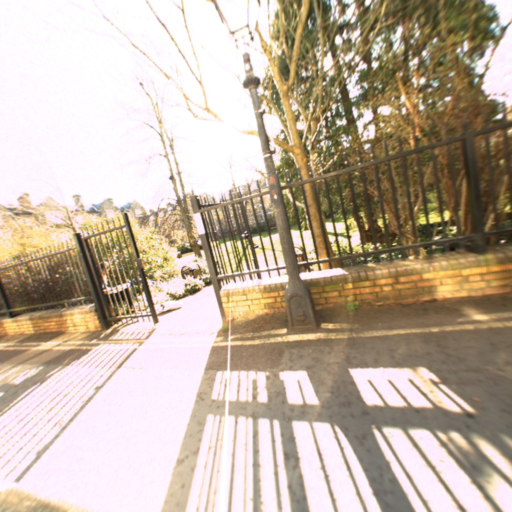}} \hspace{0.05cm}
\subfloat{\includegraphics[width=0.15\linewidth]{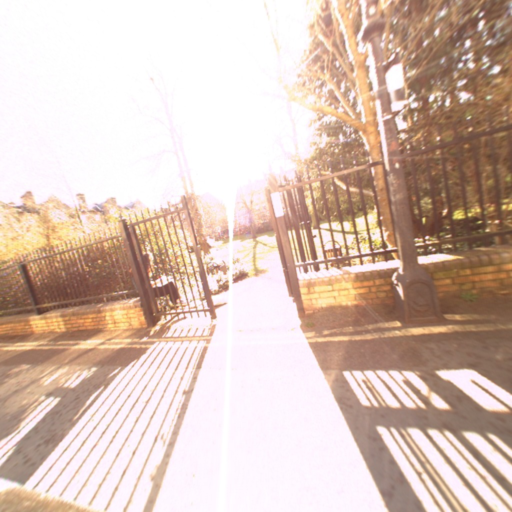}}
\hspace{-0.1cm}
\subfloat{\includegraphics[width=0.15\linewidth]{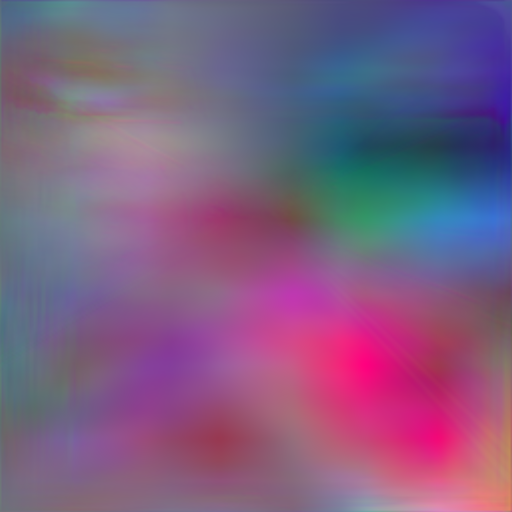}} \hspace{0.05cm}
\subfloat{\includegraphics[width=0.15\linewidth]{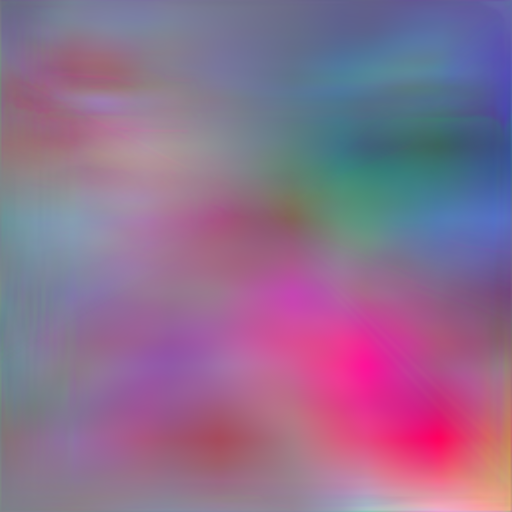}} \hspace{0.05cm}
\subfloat{\includegraphics[width=0.15\linewidth]{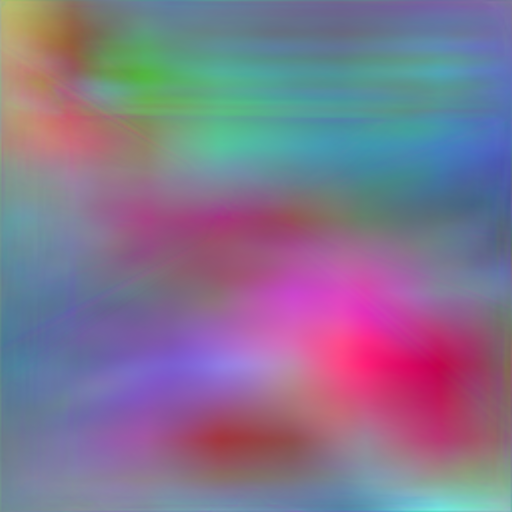}} \hspace{0.05cm}
\subfloat{\includegraphics[width=0.15\linewidth]{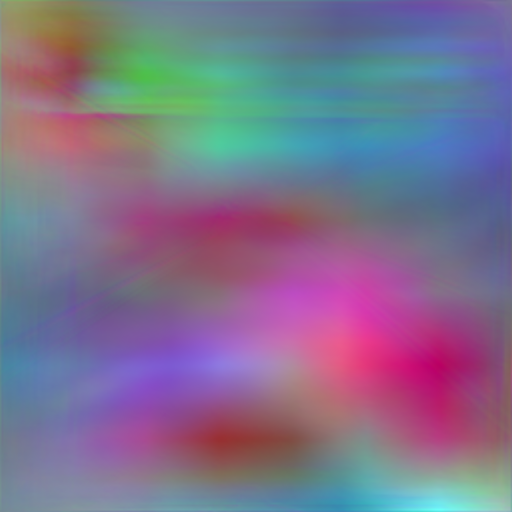}} \hspace{0.05cm}
\subfloat{\includegraphics[width=0.15\linewidth]{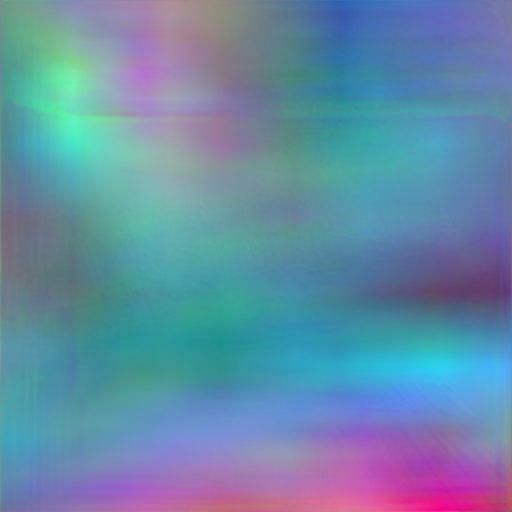}} \hspace{0.05cm}
\subfloat{\includegraphics[width=0.15\linewidth]{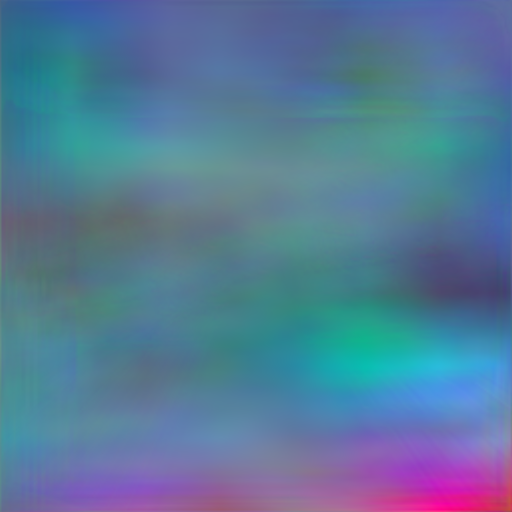}}
\caption{Learnt feature visualizations from PCA dimensionality reduction. Despite drastic appearance changes, the network is capable of correctly identifying the similarities between the anchor and positive pair, whilst discriminating the negative pair.\looseness=-1}
\label{fig: viz}
\end{figure*}

\textbf{Dataset.}
To train the proposed features we make use of the \emph{RobotCar Seasons} dataset \cite{Sattler2018a}.
This is a subset of the original RobotCar dataset \cite{Maddern} focused on cross-seasonal revisitations.
It provides a set of consecutive frames at 49 unique locations, each at different times of the year including sun, rain, dawn, overcast, dusk and night.
Additionally, a reference pointcloud and poses are provided. 
However, it it still not possible to obtain accurate cross-seasonal pixel-level matches due to pose inconsistency.
Fortunately, our system does not require this type of correspondence supervision. 

The dataset is split into a training and validation set of 40 and 9 locations, respectively.
The training triplets are generated on the fly. 
From an \textbf{A}nchor image at a given season and location, a random target season is selected for the \textbf{P}ositive sample.
The closest frame within that season is found by calculating the distance between the respective GPS readings.
Finally, the \textbf{N}egative sample is obtained by randomly sampling from a different RobotCar Seasons location, without any restriction on the season.

\textbf{Training.}
Using this data, Deja-Vu is trained for 160 epochs with a base learning rate of 0.001 and an SGD optimizer. 
The contextual triplet loss margin was typically fixed to $m = 0.5$ since the similarity between images is constrained to the range [0, 1].
In order to provide a more compact representation, the dimensionality of the features was restricted to $n=10$, with a consistency loss weight $\alpha = [0, 1]$.

\textbf{Feature visualization.}
In order to visualize the $n$-dimensional features produced by the network, we apply PCA and map the features to the RGB cube.
Three pairs of examples are shown in shown in Figure \ref{fig: viz}.
This triplet helps to illustrate some of the most challenging aspects of the task at hand. 
This includes the drastic appearance changes between different times of day, night-time motion blur from the increased exposure and sunburst/reflections.
Despite this, the feature maps for the anchor and positive appear globally similar and distinct to the negative pair.

\begin{table}[b]
\centering
\begin{tabular}{|c||c|c|}
\hline
Features & Seasonal AUC & Cross-season AUC\\ 
\hline\hline
SIFT \cite{Lowe2004} & 80.78 & 46.79 \\
RootSIFT \cite{Arandjelovic2012} & \textbf{97.15} & \underline{59.75} \\
ORB \cite{Rublee2011} & \underline{96.60} & \textbf{66.99} \\
\hline
SIFT + CX & 94.42 & 64.58 \\
RootSIFT  + CX & 95.55 & 68.36 \\
ORB + CX & 96.26 & 70.54 \\
VGG \cite{Simonyan2015} + CX & 99.05 & 73.03 \\
NC-Net \cite{Rocco2018} + CX & 97.58 & 74.03 \\ 
D2-Net \cite{Dusmanu2019} + CX & 98.70 & \underline{74.96} \\
SAND \cite{Spencer2019} + CX & \textbf{99.74} & 74.86 \\
NetVLAD \cite{Arandjelovi} + CX & \underline{99.41} & \textbf{77.57} \\
\hline
\ac{DVF} - $\alpha=0$ & 99.30 & 93.82 \\
\ac{DVF} - $\alpha=0.2$ & \textbf{99.82} & \textbf{96.56} \\
\ac{DVF} - $\alpha=0.4$ & 99.59 & 91.37 \\
\ac{DVF} - $\alpha=0.6$ & \underline{99.76} & 93.46 \\
\ac{DVF} - $\alpha=0.8$ & 99.52 & \underline{94.12} \\
\ac{DVF} - $\alpha=1$ & 99.47 & 92.94 \\
\hline
\end{tabular}
\caption{Within season and cross-season AUCs from Figure \ref{fig: conf_mats}. Incorporating the contextual similarity improves performance on hand crafted baselines. The proposed approach further increases accuracy by a large margin.}
\label{table: x_season}
\end{table}

\begin{figure*}
\centering
\subfloat[ORB]{\hspace{-0.4cm} \includegraphics[width=0.5\linewidth]{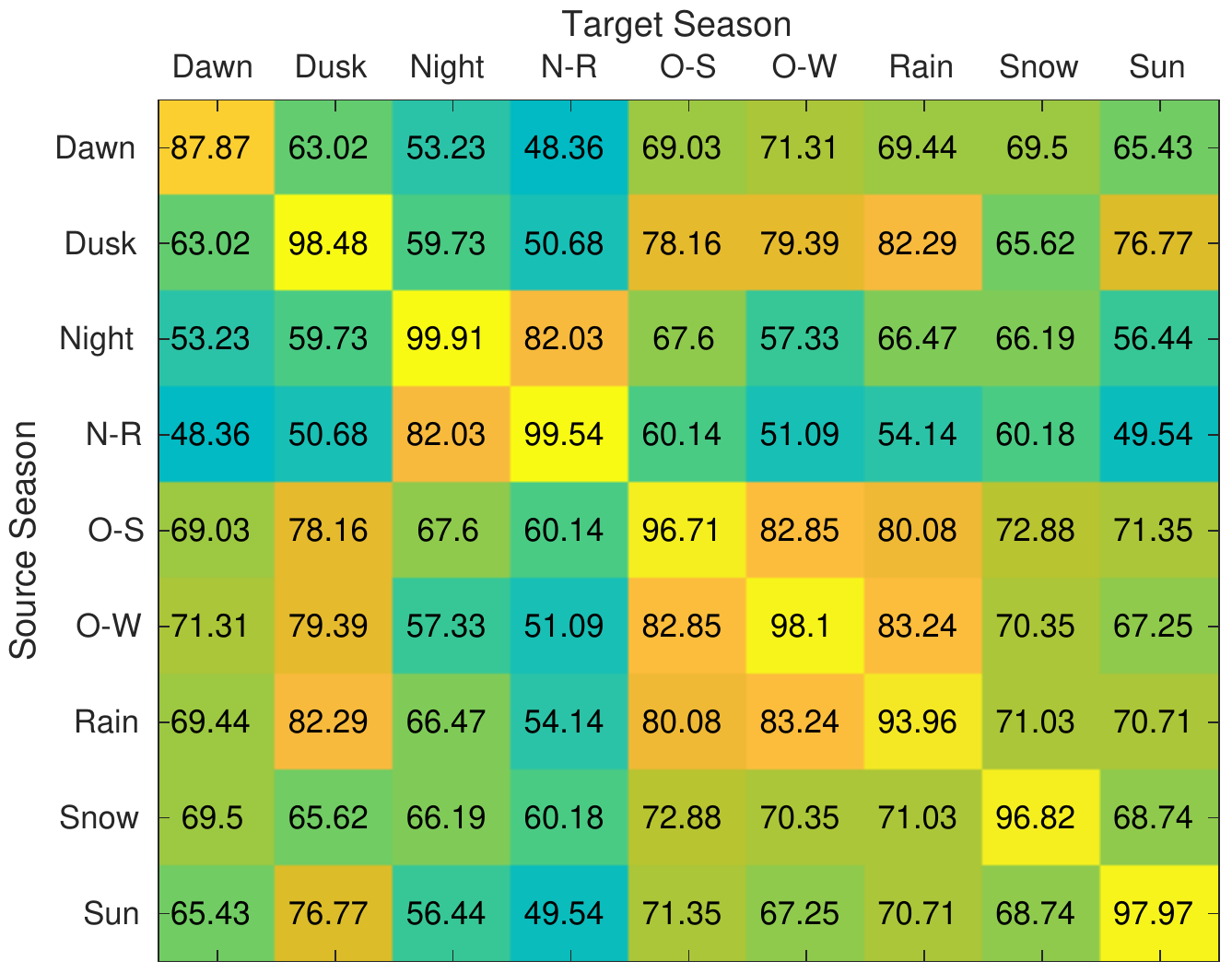}} \hspace*{0.2cm}
\subfloat[SAND + CX]{\includegraphics[width=0.5\linewidth]{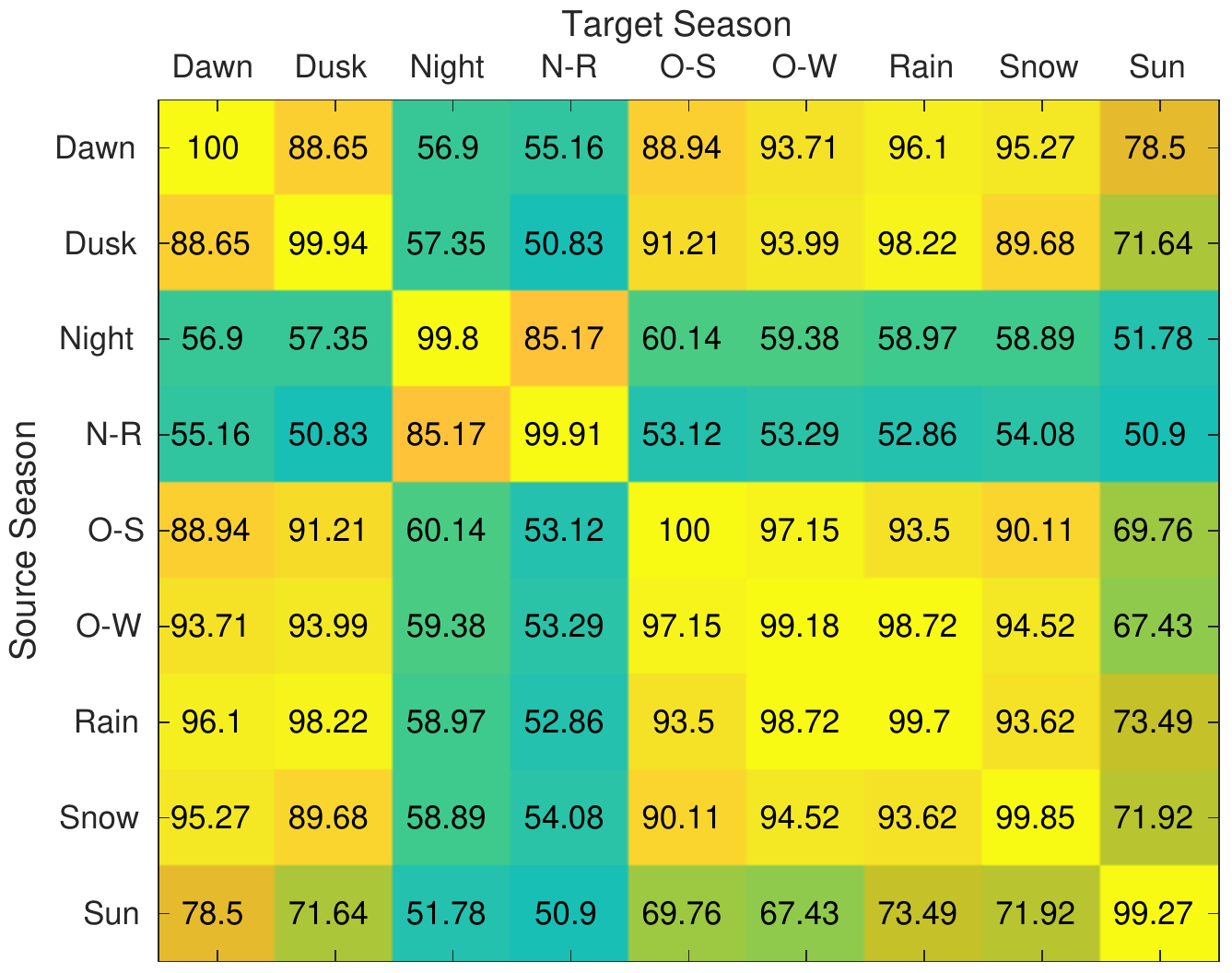}}
\\
\subfloat[NetVLAD + CX]{\hspace{-0.4cm} \includegraphics[width=0.5\linewidth]{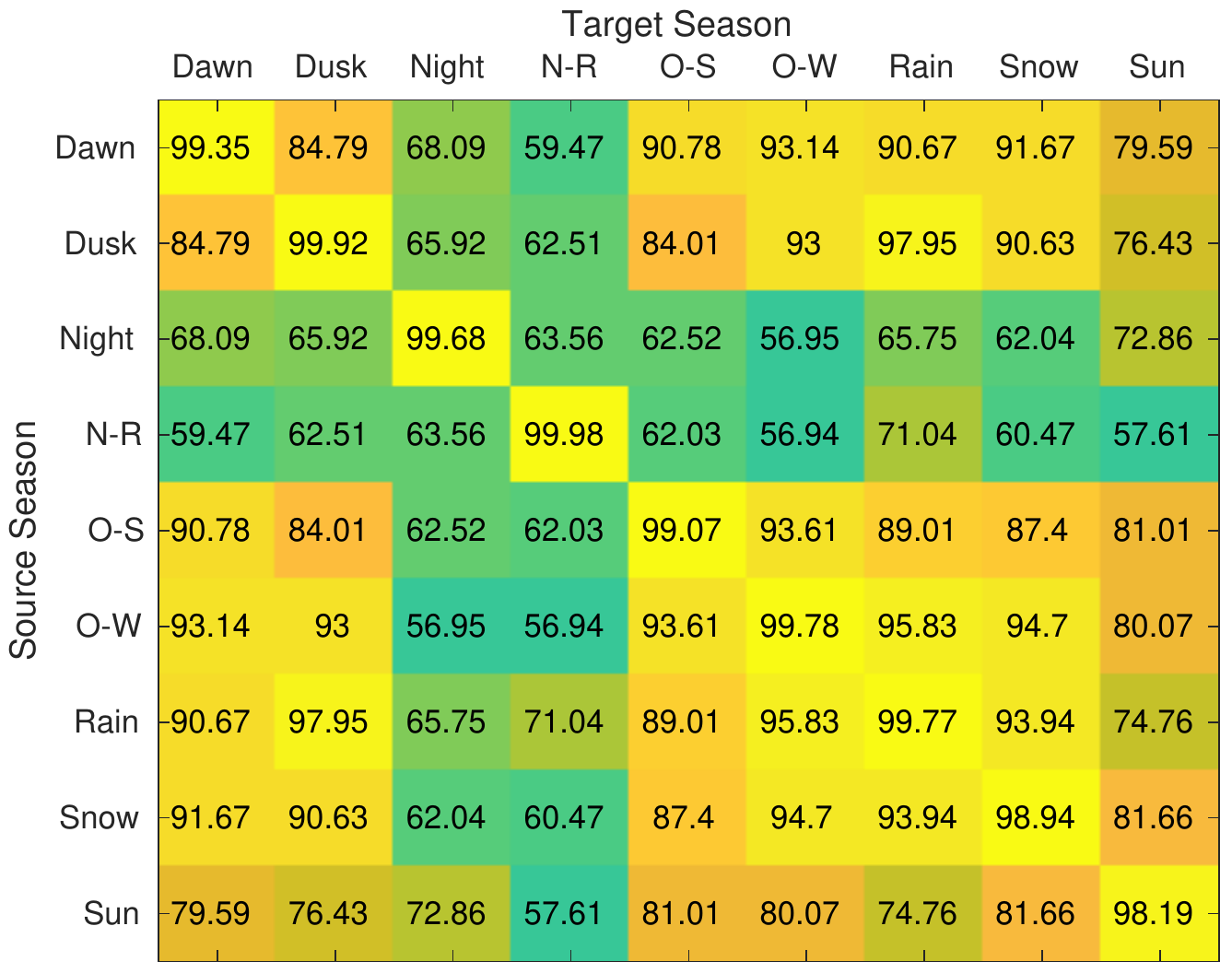}} \hspace*{0.2cm}
\subfloat[\ac{DVF} (Proposed)]{\includegraphics[width=0.5\linewidth]{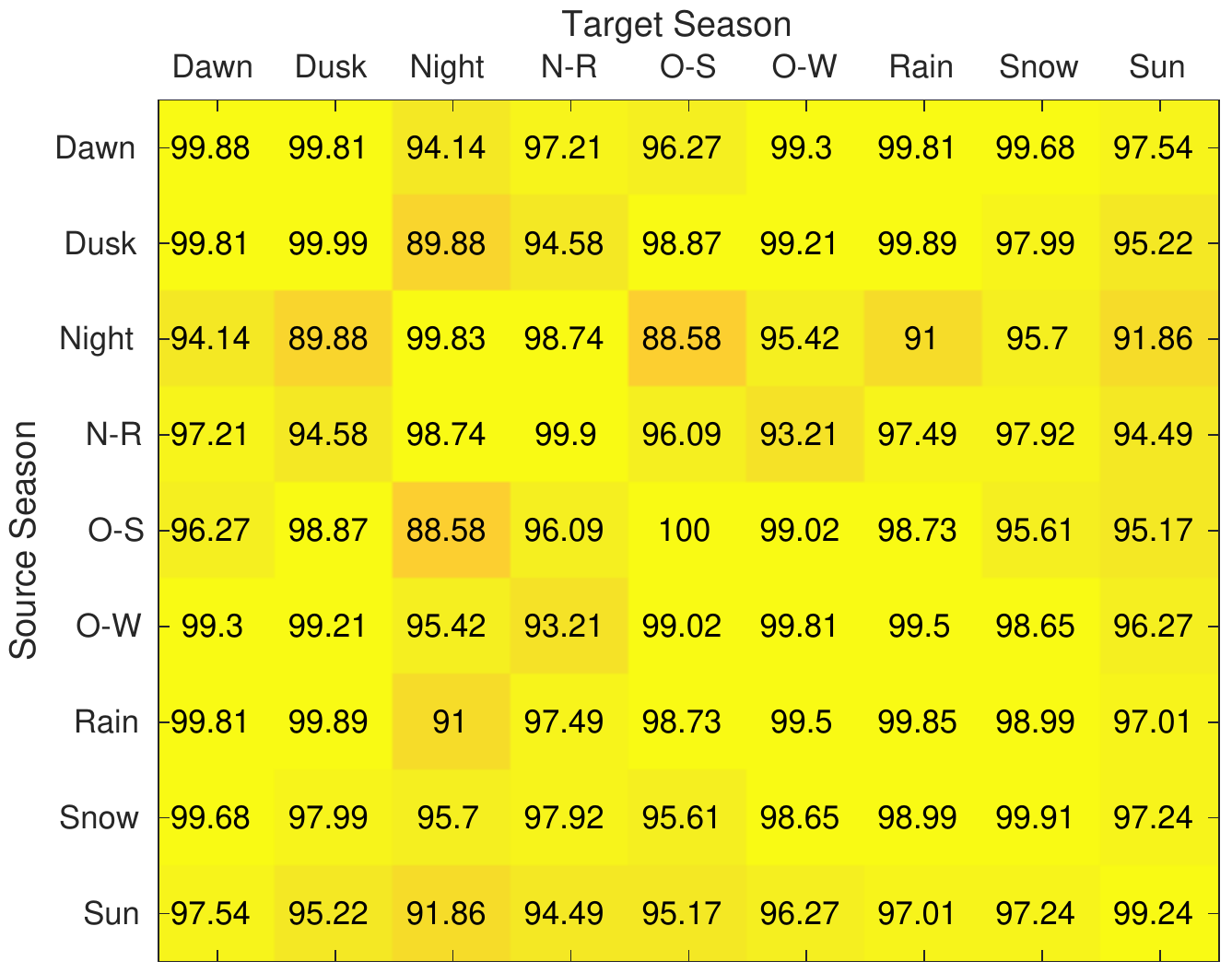}}
\caption{Performance matrices for baselines and the proposed \ac{DVF}. The diagonals represent localization within each season, whereas all other cells perform cross-season localization. For summarized results see Table \ref{table: x_season}. N-R: Night-Rain, O-S: Overcast-Summer, O-W: Overcast-Winter.}
\label{fig: conf_mats}
\end{figure*}

\textbf{Cross-seasonal AUC.}
The baselines and proposed \ac{DVF} are evaluated based on the \ac{AUC} of the ROC curve when classifying a pair of images as corresponding to the same location or not. 
In this context, we consider all images within each RobotCar Seasons location as ``true positives'', regardless of the season and their exact alignment.

These results are shown in Figure \ref{fig: conf_mats} as a series of performance matrices indicating the classification performance between all possible season combinations. 
The diagonal corresponds to classification within each season, whereas all other blocks represent cross-seasonal classification. 
This is summarized in Table \ref{table: x_season}, where we show that the proposed features outperform all baselines.
These baselines were obtained by using the code and features provided by the corresponding authors/libraries. 
Additionally, we show how the used similarity metric can improve performance even in traditional methods using ORB, SIFT and RootSIFT.

\begin{figure*}[!ht]
\centering
\makebox[20pt]{\raisebox{30pt}{\rotatebox[origin=c]{90}{DVF}}}%
\stackon[-5pt]{\subfloat{\includegraphics[width=0.3\linewidth]{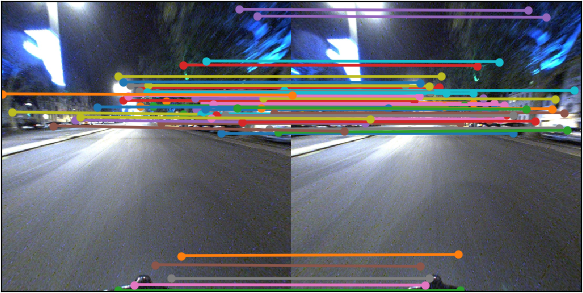}} \hspace*{0.2cm}}{\hspace{-10pt}Night}
\stackon[-5pt]{\subfloat{\includegraphics[width=0.3\linewidth]{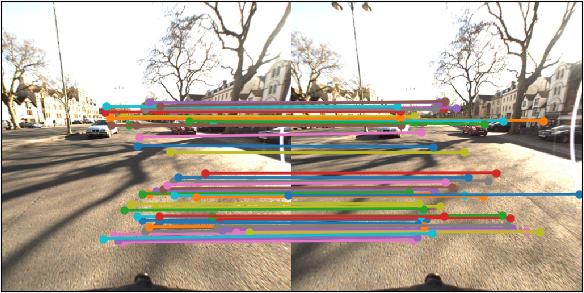}} \hspace*{0.2cm}}{\hspace{-10pt}Dawn}
\stackon[-5pt]{\subfloat{\includegraphics[width=0.3\linewidth]{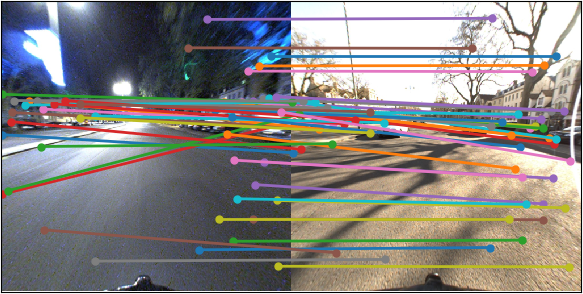}}}{Cross-season}
\\
\makebox[20pt]{\raisebox{30pt}{\rotatebox[origin=c]{90}{SAND}}}%
\subfloat{\includegraphics[width=0.3\linewidth]{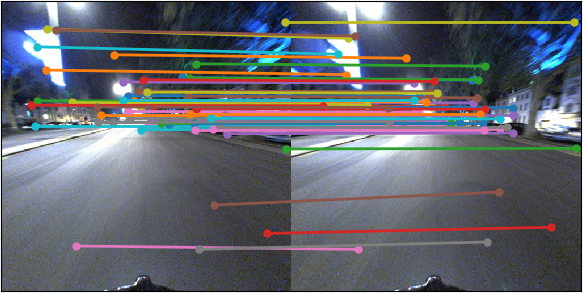}} \hspace*{0.2cm}
\subfloat{\includegraphics[width=0.3\linewidth]{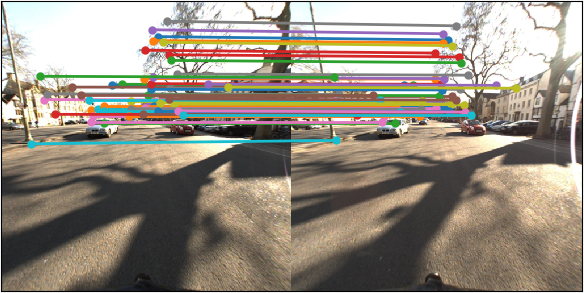}} \hspace*{0.2cm}
\subfloat{\includegraphics[width=0.3\linewidth]{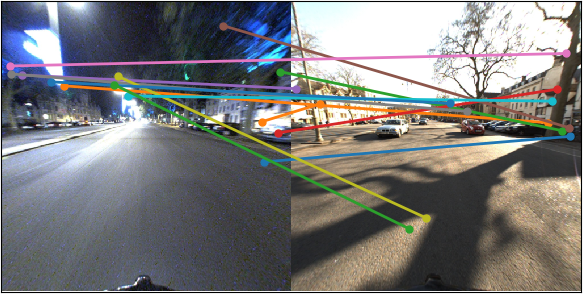}}
\\
\makebox[20pt]{\raisebox{30pt}{\rotatebox[origin=c]{90}{D2-Net}}}%
\subfloat{\includegraphics[width=0.3\linewidth]{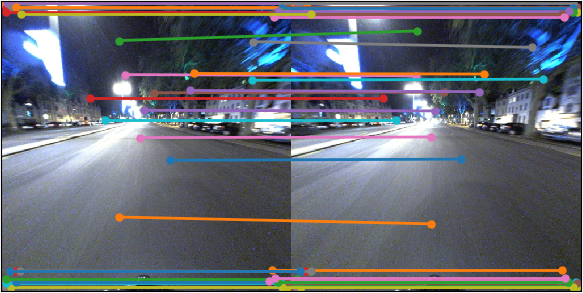}} \hspace*{0.2cm}
\subfloat{\includegraphics[width=0.3\linewidth]{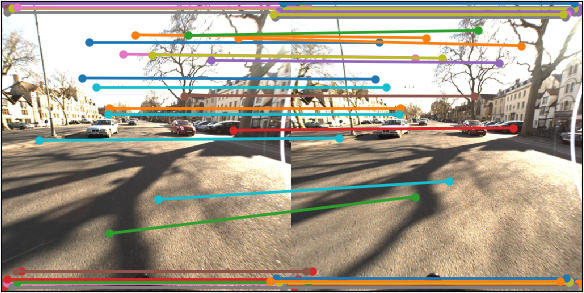}} \hspace*{0.2cm}
\subfloat{\includegraphics[width=0.3\linewidth]{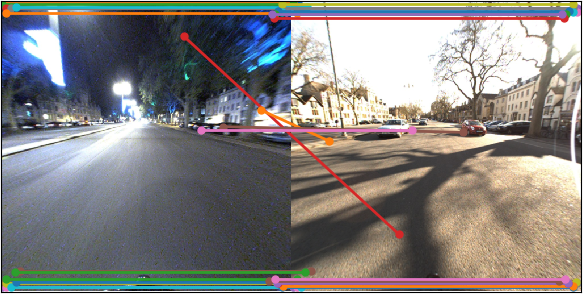}}
\caption{Sample seasonal and cross-seasonal matches obtained by DVF (proposed), SAND and D2-Net, respectively. However, they fail when attempting to match scenes with drastic appearance changes.}
\label{fig: s-matches}
\end{figure*}

\textbf{Sparse feature matching.}
Despite producing primarily a dense feature map representation, Deja-Vu can also be used to perform sparse cross-seasonal matching. 
This is worthy of note, given that the proposed method does not make use of any spatial information when training.
The system is only required to produce globally ``\emph{similar}'' or ``\emph{dissimilar}'' feature maps, with no context on what regions of the images match to each other.

Recently, a new dataset \cite{Larsson2019} was proposed containing cross-seasonal correspondences. 
However, additional experiments in the supplementary material show that this dataset is still not accurate enough to provide meaningful evaluation data, especially in the case of the RobotCar Seasons dataset.
As such, we provide quantitative results on \cite{Larsson2019} as supplementary material, and instead show qualitative performance compared to two recent state-of-the-art feature representations, SAND \cite{Spencer2019} and D2-Net \cite{Dusmanu2019}, using the models provided by the respective authors.

In order to provide the keypoint locations at which to match, we use the well established Shi-Tomasi corner detector \cite{JianboShi1994}.
In the case of D2-Net we use their own provided keypoint detection module.
The detected descriptors are matched using traditional techniques, such as mutual nearest neighbour and the ratio test, and refined using RANSAC \cite{Foley1981}.
In all images we show a representative subset of the obtained inliers to avoid cluttering the visualizations.

The first two columns in Figure \ref{fig: s-matches} represent short-term matches between consecutive frames. 
Here it can be seen how all methods perform well, obtaining multiple matches.
However, in the case where we try to perform matching between two different seasons at different times, \ie the final column, performance drops significantly for SAND and D2-Net. 
Meanwhile, \ac{DVF} is still capable of handling drastic changes in appearance.

\textbf{Cross-Seasonal Relocalization.}
Finally, we show how Deja-Vu can be used to perform 6-DOF cross-seasonal relocalization.
In practice this means that localization can be performed in previously unseen conditions without requiring additional training or fine-tuning.
In order to demonstrate this, PoseNet \cite{Kendall2015} is trained on a subset of RobotCar sequences from one season and evaluated on a corresponding subset from a different season.

The baseline is obtained by training PoseNet in a traditional manner.
Meanwhile, all other feature variants are incorporated by replacing the input image to the network with its corresponding dense $n$-dimensional feature representation, namely D2-Net, SAND and the proposed \ac{DVF}.
These features correspond to those in Table \ref{table: x_season}, which are left fixed during PoseNet training.

\begin{figure}
\vspace*{-0.5cm}
\centering
\includegraphics[width=\linewidth]{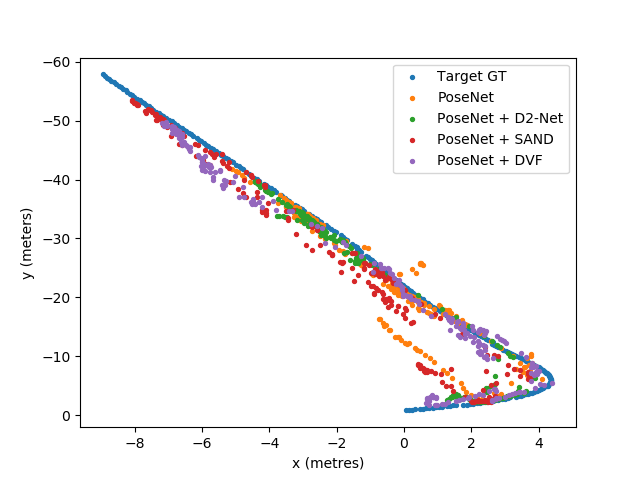}
\caption{Predicted localization using PoseNet models trained under different seasonal conditions. The variant using the proposed Deja-Vu features follows the target trajectory more closely.}
\label{fig: poses}
\end{figure}

From the results in Table \ref{table: posenet}, it can be seen how the \ac{DVF} variants clearly outperform the baselines, with the best one almost halving the error.
Figure \ref{fig: poses} shows some qualitative results from the localization pipeline.
As expected, the proposed PoseNet variant using Deja-Vu features follows the ground truth poses more closely, despite having been trained on different weather conditions.

\begin{table}[!h]
\centering
\begin{tabular}{ |c||c|c|}
\hline
Method & \textbf{P} (m) & \textbf{R} (deg/m) \\
\hline
\hline
PoseNet \cite{Kendall2015} &  10.3459 &   0.0170 \\
D2-Net \cite{Dusmanu2019} &  11.1858 &   \textbf{0.0029} \\
SAND \cite{Spencer2019} &   7.3386 &   0.0045 \\
\hline
DVF - $\alpha = 0$ & \textbf{5.5759} &   0.0050 \\
DVF - $\alpha = 1$ & \underline{7.2076} &   \underline{0.0036} \\
\hline
\end{tabular}
\caption{\textbf{P}osition (meters) and \textbf{R}otation (deg/meter) error when localizing in a RobotCar sequence of one season using a sequence of a different season.}
\label{table: posenet}
\end{table}

\section{Conclusions \& Future Work}
In this paper we have proposed Deja-Vu features, a novel approach to dense feature learning which is robust to temporal changes.
We have achieved this in a largely unsupervised manner, removing the need for exact pixel-wise matches between cross-season sequences.
In combination with the relational nature of the supervision, this can generate much larger amounts of training data by simply using rough alignment obtained automatically from GPS. 

We have shown how the use of contextual similarity can improve relocalization performance, even in well established methods using hand-crafted features.
While state-of-the-art same season localization methods tend to perform with high accuracy, their cross-seasonal performance is not comparable. 
On the other hand, Deja-Vu has over 90\% accuracy and can still perform pixel-level matching between complex seasons.

We hope this is a step towards generalizing feature representation in complex tasks and environments.
Interesting avenues for future work include introducing some level of spatial constraints into the proposed loss metrics.

\subsection*{Acknowledgements}
This work was funded by the EPSRC under grant agreement (EP/R512217/1). 
We would also like to thank NVIDIA Corporation for their Titan Xp GPU grant.
{\small
\bibliographystyle{ieee}
\bibliography{CVPR2020}}
\end{document}